\pdfoutput=1

\documentclass[11pt]{article}
\usepackage[final]{acl}

\usepackage{times}
\usepackage{latexsym}

\usepackage[T1]{fontenc}

\usepackage[utf8]{inputenc}
\usepackage{amsmath}
\usepackage{tcolorbox}
\usepackage{multirow}
\usepackage{array}
\usepackage{booktabs}
\usepackage{graphicx}
\usepackage{booktabs}
\usepackage{colortbl}
\usepackage{tabularx}
\usepackage{array}
\usepackage[dvipsnames]{xcolor}
\usepackage{enumitem}
\usepackage{adjustbox}

\providecommand{\cite}[1]{\cite{#1}}
\usepackage{microtype}

\usepackage{inconsolata}

\usepackage{graphicx}

\title{\textsc{MedFact}: A Large-scale Chinese Dataset for Evidence-based Medical Fact-checking of LLM Responses}

\author{
    Tong Chen$^{1,3,}$\thanks{Equal contribution.}, Zimu Wang$^{2,3,*}$, Yiyi Miao$^{1,3}$, Haoran Luo$^1$, Yuanfei Sun$^1$, \\
    \textbf{Wei Wang$^2$, Zhengyong Jiang$^{1,}$\footnotemark[2], Procheta Sen$^{3,}$\footnotemark[2], Jionglong Su$^{1,}$\thanks{Corresponding authors.}} \\
    $^1$School of AI and Advanced Computing, Xi'an Jiaotong-Liverpool University, China \\
    $^2$School of Advanced Technology, Xi'an Jiaotong-Liverpool University, China \\
    $^3$Department of Computer Science, University of Liverpool, United Kingdom \\
    \texttt{\{Tong.Chen19,Zimu.Wang19\}@student.xjtlu.edu.cn} \\
    \texttt{\{Zhengyong.Jiang02,Jionglong.Su\}@xjtlu.edu.cn, Procheta.Sen@liverpool.ac.uk}
}

\begin{document}
\maketitle
\begin{abstract}
Medical fact-checking has become increasingly critical as more individuals seek medical information online. However, existing datasets predominantly focus on human-generated content, leaving the verification of content generated by large language models (LLMs) relatively unexplored. To address this gap, we introduce \textsc{MedFact}, the first evidence-based Chinese medical fact-checking dataset of LLM-generated medical content. It consists of $1,321$ questions and $7,409$ claims, mirroring the complexities of real-world medical scenarios. We conduct comprehensive experiments in both in-context learning (ICL) and fine-tuning settings, showcasing the capability and challenges of current LLMs on this task, accompanied by an in-depth error analysis to point out key directions for future research. Our dataset is publicly available at \url{https://github.com/AshleyChenNLP/MedFact}.
\end{abstract}

\section{Introduction}

Nowadays, over one-third of American adults have sought medical information online before consulting a healthcare professional \cite{article}. However, the intentional proliferation of medical misinformation presents substantial risks to public health. As a result, medical fact-checking has emerged as a critical task to verify the authenticity of online medical content. This process involves assessing both the medical claims and the supportive or refuted evidence to enhance transparency in medical information and mitigate the spread of misinformation \cite{zhao-etal-2024-pacar}.

Existing medical fact-checking datasets have primarily focused on human-generated content.
With the advent of large language models (LLMs) and their growing use in medical counseling \cite{10.1145/3702647,wang-etal-2025-posts,na-etal-2025-survey}, these datasets are not equipped to address the distinct challenges posed by LLM-generated content, where the embedded parametric knowledge often lacks clear evidence or precise medical details, resulting in the hallucination issue and factual inaccuracies \cite{peng2023doesincontextlearningfall,NEURIPS2024_e560a0b2}.
Moreover, these datasets remain limited in scale, making it challenging to effectively train and evaluate LLMs for medical fact-checking tasks, with only $300$ and $750$ samples in CoVERT \cite{mohr-etal-2022-covert} and HealthFC \cite{vladika-etal-2024-healthfc}, respectively.
Overall, the limited adaptability and scale of existing datasets restrict their practicality, highlighting a significant research gap in this area.

To tackle these challenges, we introduce \textsc{MedFact}, the first evidence-based Chinese medical fact-checking dataset for medical content generated by LLMs.
As shown in Figure \ref{fig:pipeline}, we begin by collecting medical questions from the webMedQA dataset \cite{he2019applying} and generating responses using LLMs. These responses are then decomposed and decontextualized into atomic claims, each of which is evaluated for check-worthiness before retrieving relevant evidence through an ``LLM-then-Human'' verification pipeline, ensuring both efficiency and quality.
Using this dataset, we conduct extensive experiments in both in-context learning (ICL, \citealp{NEURIPS2020_1457c0d6}) and fine-tuning settings, demonstrating the effectiveness of large-scale models in leveraging parametric knowledge and the adaptability of smaller models for this task.
Our findings also reveal persistent reasoning challenges, underscoring the unique difficulty and the urgent need for specialized methodologies in this domain.

Our main contributions are as follows: (1) We introduce \textsc{MedFact}, the first evidence-based Chinese dataset targeting LLM-generated medical content. (2) We conduct extensive experiments to showcase existing LLMs on this task and highlight the challenges by reasoning-oriented models. (3) We present a thorough error analysis to identify key areas for future research, including handling medical ambiguity, recognizing semantic containment, and understanding medical synonymy.

\section{Related Work}

\paragraph{Medical Fact-checking.} Existing medical fact-checking datasets primarily focus on content generated by humans.
SciFact \cite{wadden-etal-2020-fact} reformulates expert-written claims in biomedical literature and pairs them with summaries that provide evidence.
HEALTHVER \cite{sarrouti-etal-2021-evidence-based} specializes in public health, particularly for verifying claims concerning health advice and treatments.
CoVERT \cite{mohr-etal-2022-covert} targets the verification of medical claims during COVID-19, identifying and validating misinformation during this global health crisis.
HealthFC \cite{vladika-etal-2024-healthfc} is a bilingual dataset focused on health-related claims, annotated by medical experts and supported by systematic reviews and clinical trials.
However, fact-checking datasets based on LLM-generated content in the medical domain remain unexplored, posing a significant research gap in the era of LLMs.

\paragraph{Fact-checking of LLM Responses.} With the increasing prevalence of LLMs, recent research has shifted to evaluate the factuality of LLM-generated context. HaluEval \cite{li-etal-2023-halueval} is designed to evaluate factuality around three tasks: knowledge-based discourse, summarization, and world knowledge question answering.
Attributable to Identified Sources (AIS, \citealp{rashkin-etal-2023-measuring}) emphasizes the factuality of dialogue systems with pre-injected background knowledge.
Under the open-domain setting, FELM \cite{FELM} centers long-form responses with fine-grained factuality annotations.
BingCheck \cite{li-etal-2024-self} utilizes human annotations within the SELF-CHECKER framework.
Factcheck-Bench \cite{wang-etal-2024-factcheck} further spans three levels of granularity.
In this paper, we inherit the emphasis on fact-checking LLM responses, but our position in the medical field remains underexplored in existing research.

\section{Dataset Construction}
\begin{figure}[t!]
    \centering
    \includegraphics[width=\linewidth]{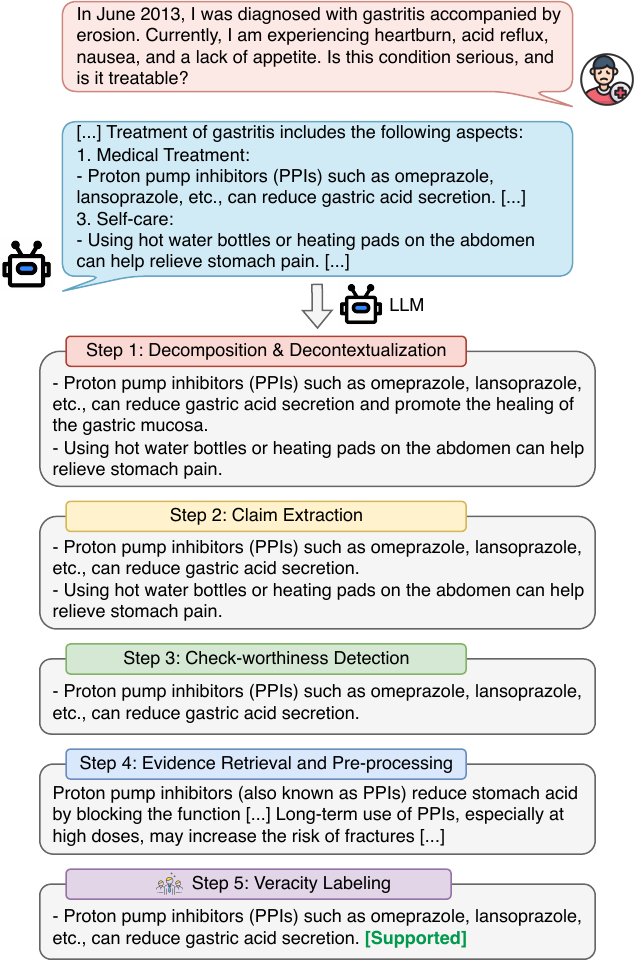}
    \caption{Overall pipeline for constructing the \textsc{MedFact} dataset.}
    \label{fig:pipeline}
    \vspace{-4mm}
\end{figure}

Figure \ref{fig:pipeline} presents the overall pipeline for constructing our \textsc{MedFact} dataset, consisting of six distinct steps. In this section, we introduce each of the steps in detail.

\subsection{Problem Definition}
The evidence-based medical fact-checking task is defined as the following: Given a set of claims $C = \{ c_1, c_2, \ldots, c_n \}$ and their evidence $E = \{ e_1, e_2, \ldots, e_n \}$, where \( c_i \) is the \( i \)-th claim and \( e_i \) is its corresponding evidence, our goal is to learn a function \( f: C \times E \to \mathcal{L} \), where $\mathcal{L}$ = $\{$\textsc{supported}, \textsc{partially supported}, \textsc{refuted}, \textsc{uncertain}, \textsc{not applicable}$\}$. For each pair \( (c_i, e_i) \), models should predict the correct label (i.e., veracity) \( y_i = f(c_i, e_i) \in \mathcal{L} \), thereby determining the degree to which the evidence supports or refutes the claim. 

\subsection{Dataset Construction}

\paragraph{Data Collection and LLM Responses Generation.}
We begin by sourcing biomedical questions from an existing dataset, webMedQA \cite{he2019applying}, from which we randomly select a subset of $1,500$ questions.
These questions span a broad range of $23$ medical topics, such as internal medicine, surgery, ophthalmology, and mental health, ensuring diversity in the represented medical information.
For each question $q_i \in Q$, we leverage Yi (\texttt{Yi-Large-Turbo}, \citealp{ai2025yiopenfoundationmodels}), an LLM with performance comparable to GPT-4 but significantly more cost-effective to generate responses $r_i = \text{LLM}(q_i)$, following prior work \cite{kang2024template}.
Here, $Q$ denotes the collection of selected questions and $r_i$ signifies the generated response corresponding to $q_i$.

\paragraph{Decomposition and Decontextualization.}
Once the questions and their responses are obtained, for each response $r_i$, we utilize DeepSeek-V2.5 \cite{deepseekai2024deepseekv2}, a model with enhanced language understanding and instruction-following capabilities, to decompose the response into a sequence of discrete statements $\{s_1, s_2, \ldots, s_p\}$, where $p$ denotes the total number of individual statements.
This process ensures that each statement is self-contained, without any irrelevant background information or extended context, thereby eliminating potential interference in subsequent analysis.

\paragraph{Claim Extraction.} 

To extract claims for verification, we reframe claim identification as a generation task rather than a classification task. Specifically, given a group of statements $\{ s_1, \ldots, s_p \}$ derived from the previous step of “Decomposition and Decontextualization,” we employ DeepSeek-V2.5 to generate a set of claims $\{ c_1, \ldots, c_m \}$, where each $c_i$ corresponds to a declarative statement that needs to be verified based on external evidence. Here, $m \leq p$, since each generated claim is derived from a specific source statement, but not all source statements necessarily yield a claim. This approach leverages the generation capabilities of LLMs to extract declarative content, in line with prior work \cite{chern2023factool, li-etal-2024-self}. We also conduct a human evaluation on a subset of $100$ questions, focusing on the steps of “Decomposition and Decontextualization” and “Claim Extraction,” with an accuracy of $100\%$ on these sampled questions, ensuring the high quality of both the dataset and all intermediate steps.

\paragraph{Check-worthiness Detection.}
Not all claims warrant verification, as some may be self-evident or trivial, while others are more significant or contentious. Therefore, we introduce a claim check-worthiness detection mechanism to identify claims that merit further verification, involving evaluating each claim $c_i$ based on four key factors: \textit{Popularity}, \textit{Public Interest}, \textit{Impact}, and \textit{Timeliness} \cite{info:doi/10.2196/19969}. The check-worthiness of a claim is framed as a binary classification task, where DeepSeek-V2.5 is prompted to assign a binary label to each claim, indicating whether it should proceed to the subsequent stages. To ensure correctness, we manually check and revise the samples according to the aforementioned criteria to determine the final set of check-worthy claims.

\begin{table}[t]
\centering
\resizebox{\linewidth}{!}{\begin{tabular}{l cccc}
\toprule
\textbf{Type} & \textbf{Train} & \textbf{Val} & \textbf{Test} & \textbf{Overall}\\
\midrule
\# of Samples & $924$ & $199$ & $198$ & $1,321$\\
\# of Claims & $5,064$ & $1,186$ & $1,159$ & $7,409$\\
\midrule
\rowcolor{Gray!10}
\multicolumn{5}{l}{\textit{Text Length (Words)}} \\
\quad Response (Avg.) & $563.14$ & $557.51$ & $569.57$ & $563.25$ \\
\quad Claim (Avg.) & $23.64$ & $23.23$ & $22.74$ & $23.43$ \\
\quad Evidence (Avg.) & $448.85$ & $439.71$ & $438.42$ & $445.75$\\
\midrule
\rowcolor{Gray!10}
\multicolumn{5}{l}{\textit{Veracity Distribution (\%)}} \\
\quad Supported & $66.86$ & $66.95$ & $64.02$ & $66.43$ \\
\quad Partially Supported & $20.14$ & $21.84$ & $22.17$ & $20.73$ \\
\quad Refuted & $0.91$ & $0.93$ & $1.29$ & $0.97$\\
\quad Uncertain & $10.31$ & $8.94$ & $10.09$ & $10.06$\\
\quad Not Applicable & $1.78$ & $1.35$ & $2.42$ & $1.81$\\
\bottomrule
\end{tabular}}
\caption{Statistics of the \textsc{MedFact} dataset.}
\vspace{-4mm}
\label{tab:dataset_stats}
\end{table}

\begin{table*}[t!]
\centering
\small
\begin{adjustbox}{width=\textwidth}
\begin{tabular}{l|c|cccc|ccccc}
\toprule
\multirow{2}{*}{\textbf{Model}} & \multirow{2}{*}{\textbf{\#Para.}} & \multicolumn{4}{c|}{\textbf{Overall Performance}} & \multicolumn{5}{c}{\textbf{Per-class F1-score}} \\ \cmidrule{3-11}
& & Accuracy & Precision & Recall & F1-score & SUP & PAR & REF & UNC & NOT \\ 
\midrule
\rowcolor{gray!10}
\multicolumn{11}{c}{\textit{Human Performance}} \\ 
\midrule
Human Performance & $-$ & $88.88$ & $74.50$ & $81.06$ & $77.02$ & $95.82$ & $78.75$ & $84.21$ & $69.65$ & $56.67$ \\
\midrule
\rowcolor{gray!10}
\multicolumn{11}{c}{\textit{In-context LLMs}} \\ 
\midrule
GPT-4o & $-$ & $67.53$ & $46.18$ & $\mathbf{52.12}$ & $46.47$ & $82.57$ & $45.07$ & $45.45$ & $40.22$ & $19.05$\\ 
GPT-4o mini & $-$ & $\mathbf{70.97}$ & $43.95$ & $45.65$ & $42.79$ & $\mathbf{84.25}$ & $45.22$ & $39.02$ & $33.33$ & $12.12$\\ 
Qwen3-30B-A3B & $30$B & $54.77$ & $41.58$ & $48.31$ & $40.26$ & $67.56$ & $52.34$ & $48.48$ & $3.51$ & $\mathbf{29.41}$\\ 
Qwen3-30B-A3B (w/ R.) & $30$B & $63.47$ & $37.63$ & $46.25$ & $37.86$ & $80.52$ & $39.53$ & $23.53$ & $20.31$ & $25.42$ \\
Qwen3-32B & $32$B & $57.71$ & $43.20$ & $47.55$ & $41.15$ & $70.27$ & $\mathbf{55.63}$ & $\mathbf{51.61}$ & $4.00$ & $24.24$\\
Qwen3-32B (w/ R.) & $32$B & $42.16$ & $37.55$ & $48.27$ & $32.34$ & $45.59$ & $47.43$ & $44.44$ & $3.85$ & $20.41$\\
GLM-4-32B & $32$B & $70.01$ & $48.45$ & $46.37$ & $\mathbf{46.81}$ & $83.05$ & $45.42$ & $42.42$ & $\mathbf{46.51}$ & $16.67$ \\
Deepseek-V2.5 & $236$B & $60.54$ & $\mathbf{50.40}$ & $37.97$ & $32.03$ & $76.43$ & $48.79$ & $11.11$ & $3.85$ & $20.00$\\
DeepSeek-V3 & $671$B & $67.66$ & $46.05$ & $39.27$ & $37.37$ & $83.21$ & $47.69$ & $25.00$ & $14.06$ & $16.90$\\ 
\midrule
\rowcolor{gray!10}
\multicolumn{11}{c}{\textit{Fine-tuned LLMs}} \\ \midrule
Qwen2.5-7B & $7$B & $68.51$ & $51.75$ & $41.08$ & $43.52$ & $81.77$ & $\mathbf{47.00}$ & $40.00$ & $48.83$ & $0.00$\\ 
Meditron3-Qwen2.5-7B & $7$B & $68.77$ & $47.67$ & $39.50$ & $41.44$ & $82.42$ & $46.43$ & $30.00$ & $48.37$ & $0.00$ \\
Qwen3-4B & $4$B & $68.51$ & $56.16$ & $41.11$ & $\mathbf{44.65}$ & $81.85$ & $46.74$ & $38.10$ & $44.79$ & $\mathbf{11.76}$ \\
Qwen3-8B & $8$B & $65.66$ & $50.20$ & $38.80$ & $41.31$ & $79.72$ & $43.11$ & $30.00$ & $48.00$ & $5.71$ \\
InternLM3-8B & $8$B & $67.73$ & $48.87$ & $\mathbf{41.69}$ & $44.03$ & $81.93$ & $46.23$ & $\mathbf{41.67}$ & $40.82$ & $9.52$\\ 
GLM-4-9B & $9$B & $\mathbf{69.20}$ & $\mathbf{57.13}$ & $39.30$ & $42.43$ & $\mathbf{82.61}$ & $45.42$ & $33.33$ & $\mathbf{50.76}$ & $0.00$\\
\bottomrule
\end{tabular}
\end{adjustbox}
\caption{Experimental results of in-context and fine-tuned LLMs on the \textsc{MedFact} dataset, in which the best performance on each type of LLM is highlighted in \textbf{bold}. SUP: Supported; PAR: Partially Supported; REF: Refuted; UNC: Uncertain; NOT: Not Applicable. ``w/ R.'' indicates that the reasoning mode is enabled.}
\vspace{-4mm}
\label{tab:results}
\end{table*}

\paragraph{Evidence Retrieval and Pre-processing.}
Afterward, we retrieve evidence from the web that may either support or refute the claims. For each check-worthy claim $c_i$, we use the Google Search API\footnote{\url{https://www.googleapis.com/}} to retrieve the top three most relevant documents. To minimize the impact of irrelevant information on the verification process, we leverage GLM-4-Long \cite{glm2024chatglm}, renowned for its long-document understanding capabilities, to extract the most pertinent and high-quality evidence sentences and consolidate them into coherent, self-contained evidence $e_i$ for each claim.

\paragraph{Veracity Labeling.}
In the final stage, we annotate the veracity $y_i$ given a claim $c_j$ and its associated evidence $e_i$ to represent the truthfulness of each claim based on the evidence, where the label is assigned in a pre-defined set including (1) \textit{Supported}: Evidence fully supports the claim; (2) \textit{Partially Supported}: Partially support with uncertainties; (3) \textit{Refuted}: Any evidence contradicts the claim; (4) \textit{Uncertain}: Relate to the claim but no sentences refute, support, or partially support it; and (5) \textit{Not Applicable}: Completely irrelevant.

To bootstrap the annotation process, we propose an ``LLM-then-Human'' approach, where we first generate preliminary labels using GLM-4-Long. These initial labels are then reviewed and refined by two trained undergraduate student annotators with medical backgrounds via an annotation platform based on Label Studio\footnote{\url{https://labelstud.io/}}. We randomly select $200$ samples from each annotator’s work and have them reviewed by a third annotator. The Inter-Annotator Agreement, measured by Cohen’s Kappa \cite{doi:10.1177/001316446002000104}, is $81.54\%$. More complete examples for the construction process are illustrated in Appendix \ref{sec:real-case}.

\subsection{Dataset Analysis}
\label{sec:analysis}

As detailed in Table~\ref{tab:dataset_stats}, the \textsc{MedFact} dataset consists of $1,321$ medical questions\footnote{The rest of the $179$ questions cannot be successfully processed because of the content moderation policy of LLMs \cite{he2024guardians}.} and $7,409$ claims. The dataset is carefully partitioned into a split of $70\%$:$15\%$:$15\%$ to ensure independence across subsets. The average text length of medical responses, claims, and evidence is $563.14$, $23.64$, and $448.85$ words, respectively.
For the veracities, the supported claims dominate ($66.43\%$), followed by partially supported ($20.73\%$), then uncertain, applicable, and refuted, reflecting the subtle nature of medical misinformation and posing significant challenges when dealing with this uneven distribution, mirroring the complexities of real-world medical scenarios. Examples of the dataset are depicted in Appendix \ref{sec:data-example}.

\section{Experiments}
\subsection{Experimental Setup}

We conduct experiments in in-context learning (ICL, \citealp{NEURIPS2020_1457c0d6}) and fine-tuning settings, with the prompt detailed in Figure \ref{fig:prompts}. Under the ICL setting, we evaluate GPT-4o (\texttt{2024-08-06}, \citealp{openai2024gpt4ocard}), GPT-4o mini (\texttt{2024-07-18}), Qwen3-30B-A3B \cite{yang2025qwen3technicalreport}, Qwen3-32B, GLM-4-32B (\texttt{0414}, \citealp{glm2024chatglmfamilylargelanguage}), DeepSeek-V2.5 \cite{deepseekai2024deepseekv2}, and DeepSeek-V3 (\texttt{0324}, \citealp{deepseekai2024deepseekv3}), where the temperature is set as $0.7$ for all models.
For fine-tuning, we focus on Qwen2.5-7B, Meditron3-Qwen2.5-7B\footnote{\url{https://huggingface.co/OpenMeditron/Meditron3-Qwen2.5-7B}}, Qwen3-4B, Qwen3-8B, InternLM3 \cite{cai2024internlm2}, and GLM-4-9B (\texttt{0414}). During fine-tuning, we set the number of epochs as $10$, the learning rate as $1e-4$, the batch size as $4$, and the number of gradient accumulation steps as $4$. Performance is assessed using accuracy, macro precision, macro recall, and macro F1-score. All experiments are conducted on $2$ NVIDIA A800 Tensor Core GPUs.
Prompts for the experiments are organized in Appendix \ref{sec:prompts}.

\subsection{Experimental Results}
Table \ref{tab:results} illustrates the results of the experimented LLMs on the \textsc{MedFact} dataset.
We also sample a subset of $2,000$ claims to verify the human performance on this task.
From the table, we have the following observations:

(1) GLM-4-32B demonstrates the best performance in the ICL setting, and Qwen3-4B achieves optimal results among the fine-tuned models. However, none of the models outperform humans, underscoring the significant challenge posed and the necessity of introducing the \textsc{MedFact} dataset. Despite the significant disparity in model sizes ($671$B for DeepSeek-V3 and $4$B for Qwen3), smaller models outperform larger ones after fine-tuning. This highlights the effectiveness of large-scale models in leveraging parametric knowledge for medical fact-checking, as well as the adaptability of smaller models to this specific task. Meditron3-Qwen2.5, specialized in clinical medicine, does not outperform Qwen2.5, which may be attributed to its improved capabilities in medical question answering, potentially at the expense of other tasks.

(2) LLMs are effective in classifying supported, followed by partial supported and refuted evidence, while the uncertain and not applicable veracities exhibit the poorest performance. After fine-tuning, the Qwen3, InternLM3, and GLM-4 models show performance gains in classifying uncertain evidence, though this is accompanied by the degradations of other evidence, highlighting the significant challenge in achieving simultaneous improvements across all evidence categories.

(3) The reasoning mode exhibit limited improvements or even decreased performance. With the reasoning mode, Qwen3-30B-A3B and Qwen3-32B showcase performance drops of $6.0\%$ and $21.4\%$ in F1-score, respectively. A notable limitation of reasoning models is their tendency to produce overly \textit{certain} outputs, like Qwen3-30B-A3B achieve a decrease of $14.4\%$ on the partially supported evidences, highlighting its inability to handle such scenarios effectively.

\subsection{Error Analysis}

To point out promising avenues for future research, we conduct a thorough analysis of errors made by LLMs and categorize them into three types:

(1) \textbf{Evidence Misunderstanding:}
LLMs often misclassify ``uncertain'' instances into other veracity categories. As shown in Table \ref{e1}, the evidence discusses the function of AST but omits the normal range, which should be classified as ``uncertain,'' but LLMs incorrectly assign ``partially supported'' to such cases, highlighting their drawbacks in handling ambiguous or incomplete evidence.

(2) \textbf{Semantic Containment Overlook:} LLMs fail to recognize the semantic containments between claims and evidence. As shown in Table \ref{e2}, the evidence exclusively supports meditation as a method for stress relief, but does not address the broader claim, resulting in the incorrect assignment of ``partially supported,'' rather than the appropriate ``supported'' classification.

(3) \textbf{Medical Synonymy Misjudgment:}
LLMs often struggle to identify semantic equivalence between the formal medical terms and their commonly used aliases. As shown in Table \ref{e3}, despite the evidence clearly describing the cyclical nature of the disease, the model incorrectly predicts a label of ``uncertain'' instead of ``supported'' because of its failure to interpret the synonymous relationship between the two terms.

\begin{table}[t!]
\centering
\footnotesize
\begin{tabular}{p{0.08\textwidth} p{0.35\textwidth}}
\toprule
\textbf{Claim} & The normal range for aspartate aminotransferase (AST) is typically between 8-40 U/L.\\
\midrule
\textbf{Evidence} & Aspartate aminotransferase (AST) is an enzyme that helps the body [...] This test is sometimes referred to as SGOT. \\ \midrule
\textbf{Prediction} & Partially Supported \\ \midrule
\textbf{Label} & Uncertain \\ \bottomrule
\end{tabular}
\caption{Example of Evidence Misunderstanding.}
\label{e1}
\end{table}

\begin{table}[t!]
\centering
\footnotesize
\begin{tabular}{p{0.08\textwidth} p{0.35\textwidth}}
\toprule
\textbf{Claim} & Daily stress can be managed through methods such as meditation, yoga, and deep breathing exercises.\\
\midrule
\textbf{Evidence} & If stress makes you feel anxious, tense, and worried, try meditation. [...] However, meditation is not a substitute for treatment. \\ \midrule
\textbf{Prediction} & Supported \\ \midrule
\textbf{Label} & Partially Supported \\ \bottomrule
\end{tabular}
\caption{Example of Semantic Containment Overlook.} 
\label{e2}
\end{table}

\begin{table}[t!]
\centering
\footnotesize
\begin{tabular}{p{0.08\textwidth} p{0.35\textwidth}}
\toprule
\textbf{Claim} & The symptoms of psoriasis may periodically worsen and improve.\\
\midrule
\textbf{Evidence} & Psoriasis is a common, chronic disease [...] Common triggers for individuals with a genetic predisposition to psoriasis [...] \\ \midrule
\textbf{Prediction} & Uncertain \\ \midrule
\textbf{Label} & Supported \\ \bottomrule
\end{tabular}
\caption{Example of Medical Synonymy Misjudgment.} 
\label{e3}
\vspace{-4mm}
\end{table}

\section{Conclusion}

We introduce \textsc{MedFact}, the first evidence-based Chinese medical fact-checking dataset for LLM-generated medical content, consisting of $1,321$ questions and $7,409$ claims, mirroring the complexities of real-world scenarios.
Experimental results in both ICL and fine-tuning settings showcase the capability and challenges of current LLMs on this task, and we perform an in-depth error analysis to point out key directions for future research.

\section*{Limitations}

Although \textsc{MedFact} pioneers the research in medical fact-checking of LLM responses, its scope is currently limited to Chinese because the medical questions are sourced from the webMedQA dataset. While this limitation does not diminish our contribution and the validity of our findings, we advocate for further research efforts to develop more diverse datasets with multilinguality.
Furthermore, similar to earlier fact-checking datasets, the label distribution of \textsc{MedFact} is imbalanced. Future work can focus on generating synthetic data or applying adversarial learning techniques to inject misinformation \cite{pan-etal-2023-attacking,10580402} to alleviate this limitation.

\section*{Ethical Considerations}

We discuss the following ethical considerations related to our \textsc{MedFact} dataset as follows: (1) \textbf{Intellectual Property.} The webMedQA dataset is distributed under the Apache-2.0 license\footnote{\url{https://www.apache.org/licenses/LICENSE-2.0}}, which is free for research use. We follow the regulations of the license and share our dataset under the same Apache-2.0 license. (2) \textbf{Annotators Treatments.} We hired student annotators and fairly pay them according to agreed salaries and workloads. (3) \textbf{Intended Use.} \textsc{MedFact} can be utilized to develop more persuasive models in the field of medical fact-checking. Researchers can also inherit our dataset design to develop their own datasets. (4) \textbf{Controlling Potential Risks.} Since the documents of \textsc{MedFact} do not contain private information and the annotation process is not necessary to make many judgments about social risks, we believe \textsc{MedFact} does not introduce any additional risks. We manually verified some randomly sampled data to ensure the dataset did not contain risky issues.

\section*{Acknowledgments}

We thank all anonymous reviewers for their valuable feedback. This research is funded by the Postgraduate Research Scholarship (PGRS) at Xi'an Jiaotong-Liverpool University, contract number FOSSP221001.



\bibliography{custom}

\appendix
\onecolumn

\section{Dataset Examples}
\label{sec:data-example}

\begin{figure}[!h]
\begin{tcolorbox}[colframe=green!50!black, colback=green!10!white, title=Supported Claim Example]
\small
\ttfamily
\textbf{Claim:} Scurvy is caused by vitamin C deficiency. \\
\textbf{Evidence:} Scurvy is rare in the United States and may occur in people with alcohol use disorders and in malnourished older adults. \textcolor{red}{Adults with vitamin C deficiency who lack vitamin C in their diets feel easily fatigued, sluggish and irritable, and may experience weight loss, muscle wasting and joint pain. After several months of vitamin C deficiency, scurvy's can develop.} A low vitamin C diet that lasts for several months can cause scurvy, which manifests itself as subcutaneous bleeding (especially around hair follicles or the appearance of bruising), bleeding gums, and bleeding in the joints [...]\\
\textbf{Veracity:} Supported
\end{tcolorbox}

\begin{tcolorbox}[colframe=yellow!45!red, colback=yellow!10!white, title=Partially Supported Claim Example]
\small
\ttfamily
\textbf{Claim:} Daily stress can be managed through methods such as meditation, yoga, and deep breathing
exercises. \\
\textbf{Evidence:} If stress makes you feel anxious, tense, and worried, try meditation. [...] \textcolor{NavyBlue}{Meditation is most commonly used for relaxation and stress relief. It is considered a beneficial complementary therapy for both the mind and body. Meditation can help you deeply relax and calm your mind.} [...] It allows you to calm the scattered thoughts that crowd your mind and cause stress. This process can improve both mental and physical health. Meditation can help you maintain a peaceful, serene, and tranquil state of mind, which is beneficial for emotional well-being and overall health. [...] However, meditation should not be used as a substitute for medical treatment. \\
\textbf{Veracity:} Partially Supported
\end{tcolorbox}

\begin{tcolorbox}[colframe=red!65!black, colback=red!10!white, title=Refuted Claim Example]
\small
\ttfamily
\textbf{Claim:} There are usually no strict restrictions on sexual activity during the recovery period after an abortion. \\
\textbf{Evidence:} \textcolor{Green}{After an abortion, it is recommended to avoid sexual intercourse for at least one month. During this period, even the use of condoms is not advisable, as the reproductive system requires time to heal.} Engaging in sexual activity too soon can lead to infections, bleeding, or even a subsequent pregnancy. [...] After one month, a follow-up examination should be conducted at the hospital to ensure the reproductive system has fully recovered, after which normal sexual activity can be resumed.
\\
\textbf{Veracity:} Refuted
\end{tcolorbox}

\begin{tcolorbox}[colframe=blue!50!black, colback=blue!10!white, title=Uncertain Claim Example]
\small
\ttfamily
\textbf{Claim:} The normal range for aspartate aminotransferase (AST) is typically between 8-40 U/L. \\
\textbf{Evidence:} Aspartate aminotransferase (AST) is an enzyme that helps the body break down amino acids. Like alanine aminotransferase (ALT), AST is usually present in low levels in the blood. Elevated AST levels may indicate liver damage, liver disease, or muscle injury. This test is sometimes referred to as SGOT (Serum Glutamic-Oxaloacetic Transaminase).
\\
\textbf{Veracity:} Uncertain
\end{tcolorbox}

\begin{tcolorbox}[colframe=blue!50!white, colback=blue!10!white, title=Not Applicable Claim Example]
\small
\ttfamily
\textbf{Claim:} Reducing stress, maintaining a healthy lifestyle, and getting adequate sleep may help improve sexual function. \\
\textbf{Evidence:} A heart-healthy lifestyle can help prevent cardiac damage that may trigger certain arrhythmias.  Reducing and managing stress, controlling high blood pressure, high cholesterol, and diabetes, and maintaining adequate sleep are essential.  The recommended sleep goal for adults is 7 to 9 hours per day.
\\
\textbf{Veracity:} Not Applicable
\end{tcolorbox}
\caption{Examples of the \textsc{MedFact} dataset with different veracity labels.}
\label{fig:examples}
\end{figure}

\newpage

\section{Prompts}
\label{sec:prompts}

\subsection{Prompts for Dataset Constuction}

\begin{figure}[!h]
\begin{tcolorbox}[colframe=Black!65!white, colback=Black!10!white, title=Prompt for Generating LLM Responses]
\small
\ttfamily
        Question: \texttt{\{given\_question\}} (Please do not need a final summary like 'To summarize', ...)
\end{tcolorbox}

\begin{tcolorbox}[colframe=Black!65!white, colback=Black!10!white, title=Prompt for Decomposition and Decontextualization]
\small
\ttfamily
        Decompose the following text into a sequence of discrete sentences. Each sentence should be self-contained and clearly express a single piece of information. Remove any irrelevant background information or extended context. The output should be in one paragraph without numbering:\\
        <in-context examples> \\
        TEXT: \texttt{\{given\_response\}}
\end{tcolorbox}

\begin{tcolorbox}[colframe=Black!65!white, colback=Black!10!white, title=Prompt for Claim Extraction]
\small
\ttfamily
        A claim is a statement that asserts something as true or false and can be verified with evidence. Your task is to accurately identify and extract every claim from the following text. Provide the extracted claim(s) without additional context or irrelevant details. If there are multiple claims, separate them clearly. Your response MUST be a list of dictionaries.  Each dictionary should contains the key 'claim', which correspond to the extracted claim. \\
        <in-context examples>\\
        TEXT: \texttt{\{processed\_response\}}
\end{tcolorbox}

\begin{tcolorbox}[colframe=Black!65!white, colback=Black!10!white, title=Prompt for Check-worthiness Detection]
\small
\ttfamily
        You are tasked with evaluating whether a claim is 'check-worthy' based on several factors. For each claim, consider the following: \\
             1. Popularity: The level of circulation or discussion of a claim. Determine whether it is commonly shared or debated online or in the media.\\
             2. Public Interest: The general public's interest in the outcome or verdict of the claim. Consider whether people would want to know if the claim is true or false.\\
             3. Impact: The potential impact of verifying or debunking the claim. Evaluate whether it would influence people's decisions, behaviors, or beliefs.\\
             4. Timeliness: The relevance of a claim to current events, trends, or discussions. Assess whether its truth or falsehood needs to be determined quickly due to its relation to ongoing topics.\\
             Given each claim, using above factors to label it as 'Yes' which means check-worthy or 'No' which means not check-worthy [no need explanation] \\
        <in-context examples>\\
        TEXT: \texttt{\{given\_claim\}}
\end{tcolorbox}

\begin{tcolorbox}[colframe=Black!65!white, colback=Black!10!white, title=Prompt for Veracity Labeling]
\small
\ttfamily
        Please determine the relationship between the following claim and evidence, and assign an appropriate label: The labels include:\\
        1. Supported: Evidence fully supports \\
             2. Partially Supported: Partial support with uncertainties \\
             3. Refuted: Evidence contradicts \\
             4. Uncertain: Evidence is insufficient or does not clearly indicate the truthfulness of the claim \\
             5. Not Applicable: Irrelevant evidence\\
        <in-context examples>\\
        Claim: \texttt{\{given\_claim\}}\\
        Evidence: \texttt{\{given\_evidence\}}\\
        Please provide the most appropriate label without giving an explanation.
\end{tcolorbox}
\caption{Prompts used for the construction of the \textsc{MedFact} dataset.}
\label{fig:construction-prompts}
\end{figure}

\subsection{Prompt for Experiments}

\begin{figure}[!h]
\begin{tcolorbox}[colframe=blue!65!green, colback=blue!10!white, title=Prompt for ICL and Fine-tuning Experimental Settings]
\small
\ttfamily
        You are a professional fact-checking assistant. Given a claim and corresponding evidence, select the appropriate label from these options: \\
             1. Supported: Evidence fully supports \\
             2. Partially Supported: Partial support with uncertainties \\
             3. Refuted: Evidence contradicts \\
             4. Uncertain: Evidence is insufficient or does not clearly indicate the truthfulness of the claim \\
             5. Not Applicable: Irrelevant evidence\\
             \\
             Respond strictly as: \\
             Judgment: <label>\\
             Claim: \texttt{\{given\_claim\}}\\
             Evidence: \texttt{\{given\_evidence\}}

\end{tcolorbox}

\caption{Prompt for in-context learning (ICL) and fine-tuning experiments.}
\label{fig:prompts}
\end{figure}

\section{Dataset Construction Example}
\label{sec:real-case}
This section presents a real case from our \textsc{MedFact} dataset. Table \ref{questionofdata} displays a question sourced from the webMedQA dataset \cite{he2019applying}, which we utilize to obtain the original response presented in Table \ref{ORes}. Following the processes of decomposition and decontextualization, we arrive at the processed response shown in Table \ref{ProR}. Tables \ref{claims1}-\ref{claims10} illustrate the final processed results, which encompass claims, corresponding evidence sources, processed evidence, machine label, and human label. These results were generated through a series of steps, including claim extraction, check-worthiness detection, evidence retrieval and pre-processing, and finally, veracity labeling.

\begin{table*}[h]
    \centering
    \small
    \begin{tabular}{|p{\dimexpr\linewidth-2\tabcolsep}|}
        \hline
        \vspace{1mm}
        What are the common symptoms of liver disease? My father usually maintains strict exercise habits and has always been in good health. He follows a regulated diet and rarely eats out. Recently, he has been experiencing mental fatigue, slight weight loss, and yellowish complexion. We are concerned about potential liver disease. \\
        \vspace{1mm} \\
        \hline
    \end{tabular}
    \caption{Question of the example from the \textsc{MedFact} dataset.}
    \label{questionofdata}
\end{table*}

\begin{table*}[h]
    \centering
    \small
    \begin{tabular}{|p{\dimexpr\linewidth-2\tabcolsep}|}
        \hline
        \vspace{1mm}
        Liver disease can manifest in various symptoms. Here are some common symptoms of liver disease: \\
        Jaundice: Liver disease may cause abnormal bilirubin metabolism, leading to yellowing of the skin and sclera. \\
        Fatigue: Impaired liver function can result in ineffective energy metabolism, causing persistent tiredness. \\
        Anorexia: As the liver participates in digestion, hepatic diseases may affect appetite and digestive functions. \\
        Weight Loss: Reduced appetite in liver disease patients often leads to significant weight reduction. \\
        Abdominal Pain: Liver disease may cause discomfort or pain in the right upper quadrant. \\
        Hepatomegaly: Hepatic enlargement may present as a palpable abdominal mass. \\
        Pruritus: Cholestasis caused by liver disease can induce skin itching. \\
        Bleeding Tendency: Impaired clotting factor synthesis may lead to gingival bleeding, epistaxis, or subcutaneous hemorrhage. \\
        Dark Urine: Urine may appear tea-colored due to liver dysfunction. \\
        Pale Stools: Stools may become clay-colored in liver disease. \\
        Hepatic Encephalopathy: Severe cases may manifest mental status changes including confusion or coma. \\
        Based on your description, your father exhibits symptoms of fatigue, weight loss, and a yellowish complexion, which may indicate liver disease, but could also be signs of other health issues. It is recommended to take your father to the hospital for a comprehensive examination as soon as possible, including liver function tests, blood tests, and ultrasound, to determine the exact cause and receive appropriate treatment. \\
        \vspace{1mm} \\
        \hline
    \end{tabular}
    \caption{Original response of the example from the \textsc{MedFact} dataset.}
    \label{ORes}
\end{table*}

\begin{table*}[h]
    \centering
    \small
    \begin{tabular}{|p{\dimexpr\linewidth-2\tabcolsep}|}
        \hline
        \vspace{1mm}
        Jaundice: Liver disease may cause abnormal bilirubin metabolism, leading to yellowing of the skin and sclera. \\
        Fatigue: Impaired liver function can result in ineffective energy metabolism, causing persistent tiredness. \\
        Anorexia: As the liver participates in digestion, hepatic diseases may affect appetite and digestive functions. \\
        Weight Loss: Reduced appetite in liver disease patients often leads to significant weight reduction. \\
        Abdominal Pain: Liver disease may cause discomfort or pain in the right upper quadrant. \\
        Hepatomegaly: Hepatic enlargement may present as a palpable abdominal mass. \\
        Pruritus: Cholestasis caused by liver disease can induce skin itching. \\
        Bleeding Tendency: Impaired clotting factor synthesis may lead to gingival bleeding, epistaxis, or subcutaneous hemorrhage. \\
        Dark Urine: Urine may appear tea-colored due to liver dysfunction. \\
        Pale Stools: Stools may become clay-colored in liver disease. \\
        Hepatic Encephalopathy: Severe cases may manifest mental status changes including confusion or coma. \\
        \vspace{1mm} \\
        \hline
    \end{tabular}
    \caption{Decomposed and decontextualized response of the example from the \textsc{MedFact} dataset.}
    \label{ProR}
\end{table*}

\begin{table*}[h]
    \centering
    \small
    \begin{tabular}{p{\dimexpr\linewidth-2\tabcolsep}}
        \toprule
        \textbf{Claim}: \textit{Liver disease may cause darkening of urine resembling tea color.} \\
        \midrule
        \textbf{Evidence Sources}:
        \begin{itemize}[noitemsep, topsep=0pt, partopsep=0pt]
            \item \textbf{Source 1}: ``Normal urine color is clear to pale yellow. However, certain factors can change the color of urine...''
            \item \textbf{Source 2}: ``The whites of eyes and skin typically appear yellow in jaundice patients due to high bilirubin levels...''
            \item \textbf{Source 3}: ``Jaundice serves as a warning sign of systemic disease, manifesting as yellow skin discoloration or scleral icterus...''
        \end{itemize}
        \textbf{Processed Evidence}: Dark or orange urine may indicate liver dysfunction, particularly when accompanied by pale stools and jaundice. \\
        \textbf{Machine Label}: Partially Supported \\
        \textbf{Label}: Uncertain \\
        \bottomrule
    \end{tabular}
    \caption{Claim for the example from the \textsc{MedFact} dataset.}
    \label{claims1}
\end{table*}

\begin{table*}[h]
    \centering
    \small
    \begin{tabular}{p{\dimexpr\linewidth-2\tabcolsep}}
        \toprule
        \textbf{Claim}: \textit{Liver disease may lead to abnormal bilirubin metabolism and yellowing of the skin and whites of the eyes.} \\
        \midrule
        \textbf{Evidence Sources}:
        \begin{itemize}[noitemsep, topsep=0pt, partopsep=0pt]
            \item \textbf{Source 1}: ``Jaundice occurs when the liver is diseased and is unable to remove bilirubin in sufficient amounts. Bilirubin is a metabolic waste product from the blood...''
            \item \textbf{Source 2}: ``The whites of the eyes and skin usually look yellow in people with jaundice. Jaundice occurs when there is a high level of bilirubin (a yellow pigment) in the blood...''
            \item \textbf{Source 3}: ``Jaundice is a warning sign of physical illness. When the skin becomes abnormally yellowish brown or the whites of the eyes turn yellow, this symptom should not be ignored...''
        \end{itemize}
        \textbf{Processed Evidence}: Jaundice is an abnormal condition of the body, mainly caused by the increase of bilirubin in the blood... \\
        \textbf{Machine Label}: Supported \\
        \textbf{Label}: Supported \\
        \bottomrule
    \end{tabular}
    \caption{Claim for the example from the \textsc{MedFact} dataset.}
    \label{claims2}
\end{table*}

\begin{table*}[h]
    \centering
    \small
    \begin{tabular}{p{\dimexpr\linewidth-2\tabcolsep}}
        \toprule
        \textbf{Claim}: \textit{Liver disease may present with lighter, off-white colored stools.} \\
        \midrule
        \textbf{Evidence Sources}:
        \begin{itemize}[noitemsep, topsep=0pt, partopsep=0pt]
            \item \textbf{Source 1}: ``Orange urine may indicate a problem with the liver or bile ducts; look for light-colored stools as well...''
            \item \textbf{Source 2}: ``Dark brown or orange urine, yellowish skin and eyes, and whitish stools may indicate liver deficiency...''
            \item \textbf{Source 3}: ``If the liver does not produce bile, or if bile is stagnant in the liver, the stools will be light-colored or white...''
        \end{itemize}
        \textbf{Processed Evidence}: Dark or orange urine may indicate liver dysfunction, particularly when accompanied by pale stools and jaundice. \\
        \textbf{Machine Label}: Supported \\
        \textbf{Label}: Supported \\
        \bottomrule
    \end{tabular}
    \caption{Claim for the example from the \textsc{MedFact} dataset.}
    \label{claims3}
\end{table*}

\begin{table*}[h]
    \centering
    \small
    \begin{tabular}{p{\dimexpr\linewidth-2\tabcolsep}}
        \toprule
        \textbf{Claim}: \textit{When liver function is impaired, the body may not be able to metabolize energy efficiently, leading to fatigue and lethargy.} \\
        \midrule
        \textbf{Evidence Sources}:
        \begin{itemize}[noitemsep, topsep=0pt, partopsep=0pt]
            \item \textbf{Source 1}: ``Energy metabolism may be affected, nutrient absorption may deteriorate, and the body may feel more and more tired and sluggish. Changes in urine and feces: When the liver is damaged, it is unable to process wastes and metabolites efficiently, leading to urination...''
            \item \textbf{Source 2}: ``If left untreated, over time, hypothyroidism can lead to other health problems, such as high cholesterol and heart problems...''
            \item \textbf{Source 3}: ``Fatigue and tiredness caused by liver cancer cannot be eliminated even if the patient lies down and rests for a long time. The main reason for fatigue is that cancer cells damage the metabolism and detoxification function of the liver...''
        \end{itemize}
        \textbf{Processed Evidence}: The liver is responsible for storing and releasing energy. When the liver is damaged, energy metabolism may be affected, nutrient absorption becomes poorer, and the body may feel more and more tired and sluggish. \\
        \textbf{Machine Label}: Supported \\
        \textbf{Label}: Supported \\
        \bottomrule
    \end{tabular}
    \caption{Claim for the example from the \textsc{MedFact} dataset.}
    \label{claims4}
\end{table*}

\begin{table*}[h]
    \centering
    \small
    \begin{tabular}{p{\dimexpr\linewidth-2\tabcolsep}}
        \toprule
        \textbf{Claim}: \textit{The liver is involved in the digestive process and liver disease may affect appetite and digestion.} \\
        \midrule
        \textbf{Evidence Sources}:
        \begin{itemize}[noitemsep, topsep=0pt, partopsep=0pt]
            \item \textbf{Source 1}: ``Statins are highly effective and safe for most patients, but some patients experience drug-related muscle pain, digestive problems, and mental fogginess. In rare cases, liver damage may result...''
            \item \textbf{Source 2}: ``The liver is involved in the metabolic processes in the body, and after suffering from hepatitis, the function of bile secretion decreases ... Gastrointestinal dysfunction and other symptoms, which in turn affects the patient's food digestion and absorption...''
            \item \textbf{Source 3}: ``The major organs of the digestive system include the liver, stomach, gallbladder, colon and small intestine...''
        \end{itemize}
        \textbf{Processed Evidence}: The liver is involved in the metabolic process in the body, after suffering from hepatitis, the function of bile secretion is reduced, which affects the digestion of fat, so there will be anorexia, gastrointestinal dysfunction, etc., which affects the patient's food digestion and absorption. \\
        \textbf{Machine Label}: Supported \\
        \textbf{Label}: Supported \\
        \bottomrule
    \end{tabular}
    \caption{Claim for the example from the \textsc{MedFact} dataset.}
    \label{claims5}
\end{table*}

\begin{table*}[h]
    \centering
    \small
    \begin{tabular}{p{\dimexpr\linewidth-2\tabcolsep}}
        \toprule
        \textbf{Claim}: \textit{People with liver disease may experience loss of appetite, which can lead to weight loss.} \\
        \midrule
        \textbf{Evidence Sources}:
        \begin{itemize}[noitemsep, topsep=0pt, partopsep=0pt]
            \item \textbf{Source 1}: ``Cirrhosis is the extensive destruction of the internal structure of the liver caused by the permanent replacement of large amounts of normal liver tissue by nonfunctional scar tissue...''
            \item \textbf{Source 2}: ``Patients with cirrhosis may experience loss of appetite, weight loss, fatigue and general malaise...''
            \item \textbf{Source 3}: ``Obesity increases your risk for diseases that can lead to cirrhosis, such as non-alcoholic fatty liver disease and...''
        \end{itemize}
        \textbf{Processed Evidence}: People with cirrhosis may experience symptoms such as loss of appetite, weight loss, fatigue and general malaise. These symptoms may be caused by impaired liver function, as the liver becomes less able to process medications, toxins, and waste products from the body... \\
        \textbf{Machine Label}: Supported \\
        \textbf{Label}: Supported \\
        \bottomrule
    \end{tabular}
    \caption{Claim for the example from the \textsc{MedFact} dataset.}
    \label{claims6}
\end{table*}

\begin{table*}[h]
    \centering
    \small
    \begin{tabular}{p{\dimexpr\linewidth-2\tabcolsep}}
        \toprule
        \textbf{Claim}: \textit{Liver disease may result in an enlarged liver that can be felt as a lump in the abdomen.} \\
        \midrule
        \textbf{Evidence Sources}:
        \begin{itemize}[noitemsep, topsep=0pt, partopsep=0pt]
            \item \textbf{Source 1}: ``There are many diseases and conditions that can damage the liver and cause cirrhosis. Some of the causes include Chronic alcoholism...''
            \item \textbf{Source 2}: ``Sometimes, liver cysts can become so large that you can feel them through your abdomen. What are the complications of liver cysts?...''
            \item \textbf{Source 3}: ``The spleen is a very small organ, usually about the size of a fist. However, many medical conditions, including liver disease and some cancers, can cause the spleen to enlarge...''
        \end{itemize}
        \textbf{Processed Evidence}: A variety of other conditions and diseases can lead to cirrhosis, including inflammation and scarring of the bile ducts, called primary sclerosing cholangitis;...Later stages may include jaundice, which is a yellowing of the eyes or skin; bleeding in the gastrointestinal tract; abdominal swelling due to fluid buildup in the abdomen; and confusion or drowsiness.... \\
        \textbf{Machine Label}: Supported \\
        \textbf{Label}: Partially Supported \\
        \bottomrule
    \end{tabular}
    \caption{Claim for the example from the \textsc{MedFact} dataset.}
    \label{claims7}
\end{table*}

\begin{table*}[h]
    \centering
    \small
    \begin{tabular}{p{\dimexpr\linewidth-2\tabcolsep}}
        \toprule
        \textbf{Claim}: \textit{Liver disease may cause discomfort or pain in the upper right abdomen.} \\
        \midrule
        \textbf{Evidence Sources}:
        \begin{itemize}[noitemsep, topsep=0pt, partopsep=0pt]
            \item \textbf{Source 1}: ``Signs of acute liver failure and may include: yellowing of the skin and eyes (jaundice) pain in the upper right abdomen abdominal bulging (ascites) nausea and vomiting general malaise...''
            \item \textbf{Source 2}: ``Pain or discomfort in the upper right region of the abdomen. Symptoms that may occur with NASH and cirrhosis (or severe scarring) include...''
            \item \textbf{Source 3}: ``Tissue samples show the presence of excess fat in the case of nonalcoholic fatty liver disease; in the case of nonalcoholic steatohepatitis...''
        \end{itemize}
        \textbf{Processed Evidence}: Signs of acute liver failure and may include: yellowing of the skin and eyes (jaundice) pain in the upper right abdomen abdominal bulging (ascites) nausea and vomiting generalized feeling of malaise (malaise) disorientation or confusion lethargy breath may have a musty or sweet taste tremor... \\
        \textbf{Machine Label}: Supported \\
        \textbf{Label}: Supported \\
        \bottomrule
    \end{tabular}
    \caption{Claim for the example from the \textsc{MedFact} dataset.}
    \label{claims8}
\end{table*}

\begin{table*}[h]
    \centering
    \small
    \begin{tabular}{p{\dimexpr\linewidth-2\tabcolsep}}
        \toprule
        \textbf{Claim}: \textit{Liver disease may lead to cholestasis, causing itchy skin.} \\
        \midrule
        \textbf{Evidence Sources}:
        \begin{itemize}[noitemsep, topsep=0pt, partopsep=0pt]
            \item \textbf{Source 1}: ``Diseases of the liver, bile ducts, or pancreas can cause cholestasis. Yellowing of the skin and sclera, itching of the skin, deepening of the color of the urine...''
            \item \textbf{Source 2}: ``Pruritus is the most common cutaneous manifestation of liver disease. In patients with liver disease, pruritus is usually associated with cholestasis, such as primary biliary cholangitis, primary sclerosing cholangitis...''
            \item \textbf{Source 3}: ``ALGS is characterized by abnormal bile duct development and involvement of extrahepatic organs (e.g., kidneys and eyes), as well as the skeletal and cardiovascular systems. 100\% of patients have liver involvement [2, 3], and in addition to jaundice, cutaneous xanthomas, and hepatomegaly, patients present with severe pruritus...''
        \end{itemize}
        \textbf{Processed Evidence}: The causes of cholestasis are divided into two categories: intrahepatic causesCauses include acute hepatitis, alcohol-related liver disease, primary biliary cholangitis (with bile duct inflammation and scarring), cirrhosis due to viral hepatitis B or C (also with bile duct inflammation and scarring), certain drugs... \\
        \textbf{Machine Label}: Supported \\
        \textbf{Label}: Supported \\
        \bottomrule
    \end{tabular}
    \caption{Claim for the example from the \textsc{MedFact} dataset.}
    \label{claims9}
\end{table*}

\begin{table*}[h]
    \centering
    \small
    \begin{tabular}{p{\dimexpr\linewidth-2\tabcolsep}}
        \toprule
        \textbf{Claim}: \textit{Liver disease may lead to decreased synthesis of clotting factors and symptoms such as bleeding gums, nosebleeds or bleeding under the skin.} \\
        \midrule
        \textbf{Evidence Sources}:
        \begin{itemize}[noitemsep, topsep=0pt, partopsep=0pt]
            \item \textbf{Source 1}: ``Bleeding gums or nose is supposed to be a common minor ailment that people mostly don't take seriously, but if you were told that it could be related to liver disease, would you still be able to relax?...''
            \item \textbf{Source 2}: ``Nosebleeds are mostly caused by inflammation of the nasal cavity, drying of the nasal mucosa and rupture of capillaries. Nosebleeds in young people may also be related to exertion, exercise and so on. These bleeding is not a big problem, timely treatment can effectively stop bleeding. However, if the bleeding is frequent, large and not easy to stop, it is not so simple, and may indicate other systemic diseases, such as liver disease, blood disease, autoimmune disease, and so on...''
            \item \textbf{Source 3}: ``Bleeding in patients with liver disease often manifests itself in a variety of ways; in addition to bleeding from the nose and gums and petechiae on the skin, there may be vomiting of blood or tarry stools...''
        \end{itemize}
        \textbf{Processed Evidence}: Why do patients with liver disease have bleeding? This is because a large amount of coagulation factors are synthesized in the liver, and after hepatocellular injury, the function of the liver to produce coagulation factors decreases, followed by a disorder of the coagulation mechanism. In cirrhosis, patients have hypersplenism and increased mechanical destruction of blood, resulting in leukopenia and thrombocytopenia... \\
        \textbf{Machine Label}: Supported \\
        \textbf{Label}: Supported \\
        \bottomrule
    \end{tabular}
    \caption{Claim for the example from the \textsc{MedFact} dataset.}
    \label{claims10}
\end{table*}

\end{document}